\documentclass[11pt]{article}
\usepackage{coling2020}
\usepackage{times}
\usepackage{url}
\usepackage{latexsym}
\usepackage{linguex}
\usepackage{graphicx}
\usepackage{booktabs}
\usepackage{subcaption}
\usepackage{tikz-dependency}
\usepackage{amsmath}

\colingfinalcopy % Uncomment this line for the final submission

\title{\textbf{Picking BERT's Brain:} Probing for Linguistic Dependencies in Contextualized Embeddings Using Representational Similarity Analysis}

\author{Michael A. Lepori \\
    Department of Computer Science\\ Johns Hopkins University\\
  {\tt mlepori19@gmail.com} \\\And
  R. Thomas McCoy \\
  Department of Cognitive Science\\ Johns Hopkins University \\
  {\tt tom.mccoy@jhu.edu} \\}

\date{}

\begin{document}
\maketitle
\begin{abstract}
As the name implies, contextualized representations of language are typically motivated by their ability to encode context. Which aspects of context are captured by such representations? We introduce an approach to address this question using Representational Similarity Analysis (RSA). As case studies, we investigate the degree to which a verb embedding encodes the verb’s subject, a pronoun embedding encodes the pronoun’s antecedent, and a full-sentence representation encodes the sentence's head word (as determined by a dependency parse).
In all cases, we show that BERT’s contextualized embeddings reflect the linguistic dependency being studied, and that BERT encodes these dependencies to a greater degree than it encodes less linguistically-salient controls. These results demonstrate the ability of our approach to adjudicate between hypotheses about which aspects of context are encoded in representations of language.
\end{abstract}

\section{Introduction}
\blfootnote{
    %
    % for review submission
    %

    %
    % % final paper: en-uk version 
    %
    % \hspace{-0.65cm}  % space normally used by the marker
    % This work is licensed under a Creative Commons 
    % Attribution 4.0 International Licence.
    % Licence details:
    % \url{http://creativecommons.org/licenses/by/4.0/}.
    % 
    % % final paper: en-us version 
    %
    \hspace{-0.65cm}  % space normally used by the marker
    This work is licensed under a Creative Commons 
    Attribution 4.0 International License.
    License details:
    \url{http://creativecommons.org/licenses/by/4.0/}.
}
Contextualized word embeddings \cite{devlin2019bert,peters2018deep}, which are vector representations of words in context, enable neural models of language to achieve dramatic performance improvements over models whose word embeddings do not have access to context \cite{pennington2014glove,mikolov2013distributed}. The most obvious explanation for the success of these models is that contextualized word embeddings can incorporate contextual information, whereas other embeddings cannot. Contextual information provides clues about the semantic and syntactic roles that a word plays in a sentence. For example, a verb might be understood differently when placed in different contexts: In sentence \ref{subjSent1}, the verb \textit{charged} means ``ran towards,'' where in sentence \ref{subjSent2} the verb means ``formally accused''.

\ex. \label{SubjVerb} \a. The bull \textbf{charged} the man. \label{subjSent1}
\b. The prosecutor \textbf{charged} the man. \label{subjSent2} 

\noindent
There are many aspects of context that could conceivably be captured in contextualized embeddings, from linear context (e.g., what word precedes this one?) to syntactic context (e.g., what is the parent of this word in a dependency tree?).
 In this work, we investigate which aspects of context are captured in these embeddings. We do so by studying the embeddings' representational geometry, which is the spatial relationship between representations of stimuli. In our case, the stimuli are contextualized embeddings of words.

We study this geometry by applying representational similarity analysis \cite[RSA]{kriegeskorte2008representational} to the contextualized word embeddings given by BERT \cite{devlin2019bert}. This allows us to ask fine-grained questions about which context words are encoded, and to what degree.\footnote{For brevity, we say that a model $M$ \textit{encodes} an aspect of context if the representational similarity between $M$'s embeddings and a hypothesis model encoding that aspect of context is greater than the representational similarity between $M$'s embeddings and a null hypothesis model. We also say that $M$ encodes a set of context words $A$ \textit{more than} another set $B$ if the representational similarity between $M$'s embeddings and the hypothesis model for $A$ is greater than the representational similarity between  $M$'s embeddings and the hypothesis model for $B$.} 

We find that the representational geometries of BERT's word and sentence embeddings reflect several linguistic dependencies. In particular, we find that BERT's embeddings of verbs encode the subject of those verbs more than they encode nouns that are not arguments of those verbs. We also find that the contextualized embeddings of pronouns encode the pronouns' antecedents more than they encode other nouns. Finally, we find that BERT's sentence embeddings encode the main verbs of sentences more than any other content words. This is consistent with the standard assumption in dependency parsing that the main verb of a sentence is the sentence's head. These results demonstrate the ability of our approach to illuminate which aspects of context are encoded in contextualized embeddings.\footnote{Code and data can be found at \url{https://github.com/mlepori1/Picking_BERTs_Brain}.}

\section{Background: Representational Geometry}
A representational geometry is the spatial arrangement of a set of vector representations. Representational geometries are typically formed by taking the pairwise dissimilarities between those representations.
For example, the representational geometry of the BERT embeddings for the subjects of a set of sentences is given by the pairwise dissimilarities between the contextualized word embeddings corresponding to those subjects.
%For example, the representational geometry of the set of BERT embeddings for subjects is given by the pairwise dissimilarities between the contextualized word embeddings corresponding to subjects in different contexts. 
When the set of representations is poorly understood, one can gain insight into it by comparing it to a set of representations that is well-understood. If the two sets have similar representational geometries, then one can infer that the sets encode similar information. Representational similarity analysis (RSA) allows for comparisons between two different representational geometries. 
In cognitive neuroscience, this technique is used to analyze distributed activity patterns in the brain (see \newcite{kriegeskorte2013representational} for a review); we use it to analyze the representations of artificial neural networks.

We compare contextualized BERT embeddings to representations that we construct to instantiate specific linguistic hypotheses, which we refer to as \textit{hypothesis models}. These models represent specific linguistic information and abstract away from all other information \cite{kriegeskorte2008representational}.
%also introduces the idea of using \textit{conceptual models} in RSA. Conceptual models implement theories that ``specify that a [source of data] represents particular information and abstracts from other information without specifying how the representation is computed". Our method utilizes conceptual models that implement specific linguistic hypotheses, and so we refer to them as hypothesis models. 
We compare the similarity among the representations of each hypothesis model to the similarity among BERT's embeddings; 
if the representational geometry of BERT's embeddings is better matched by the representational geometry of hypothesis model $A$ than by that of hypothesis model $B$,
%if the representational geometry of hypothesis model $A$ is a better match to that of BERT's embeddings than the representational geometry of hypothesis model $B$, 
we conclude that the hypothesis instantiated by $A$ is a better description of the content of BERT's embeddings. 
Each hypothesis model represents a specific type of context word (e.g., a verb's subject), while ignoring other context words.
%Each hypothesis model represents a specific set of context words, while ignoring other context words. 
Such a hypothesis model instantiates the hypothesis that this aspect of context is the only aspect represented in BERT's embeddings.
Of course, such extreme hypotheses are almost certainly wrong; our goal is not to find a perfect hypothesis but rather to find which of two hypotheses 
%provides a better description of the information encoded in these embeddings.
is closer to the truth.
%Therefore, they instantiate the hypotheses that the only aspect of context represented in BERT's contextualized embeddings 
%only represent contextual information about specific sets of context words.

\begin{figure}[h!]
 \centering
    \includegraphics[width=12cm]{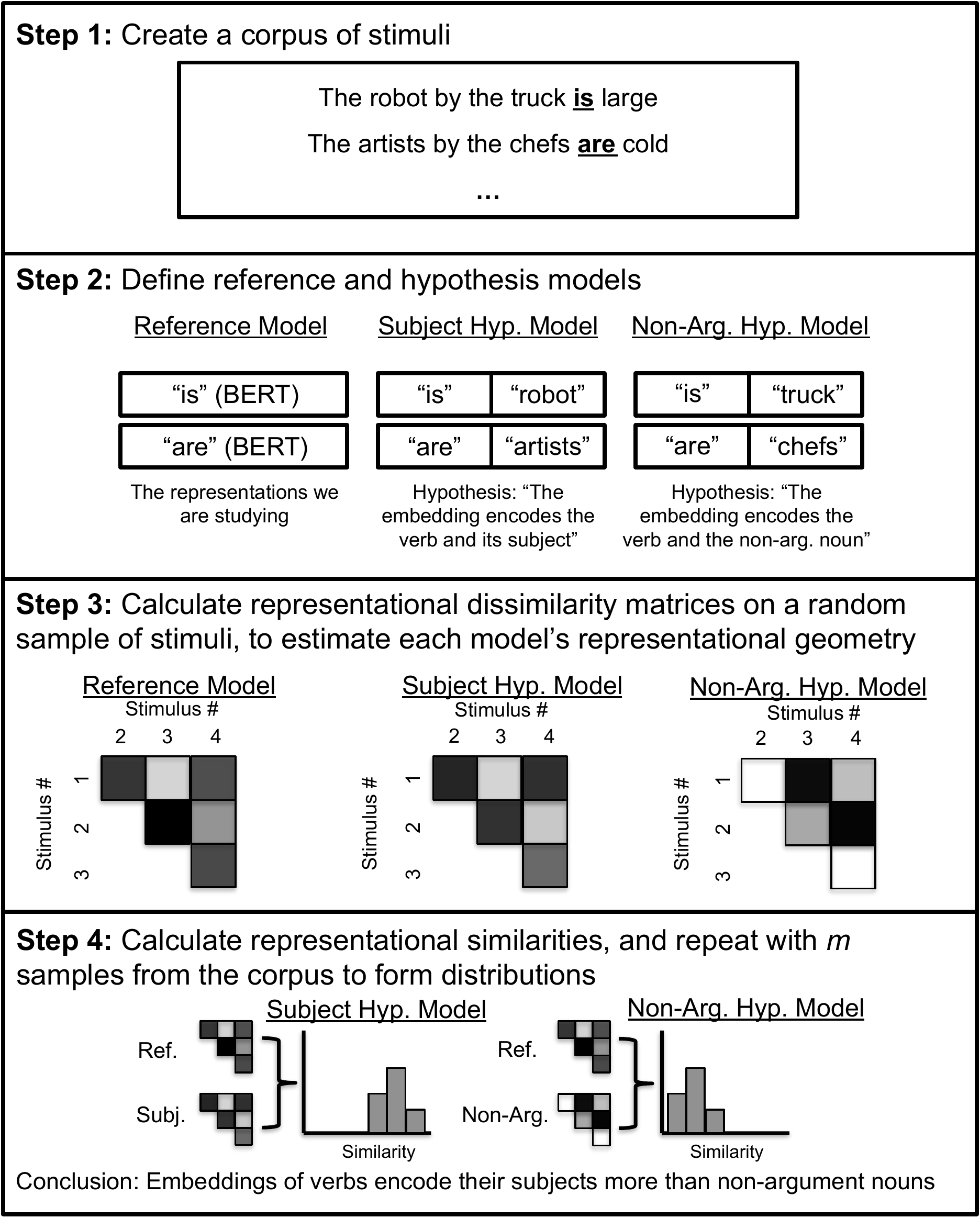}
    \caption{A summary of our approach. From this example, we would conclude that the contextualized embeddings of verbs encode their subjects more than they encode non-argument nouns.}
    \label{fig:summaryfig}
\end{figure}

\section{Probing Contextualized Embeddings} \label{RSAdesc}

Our approach works as follows: First, we create a corpus $C$ of $N$ sentences that contain the syntactic structures we wish to study (Figure \ref{fig:summaryfig}, Step 1). Next, we define a reference model $M_{Ref}$, which consists of the representations that are being investigated (e.g., the embeddings of the main verbs from every sentence in $C$).
%Next, we define a reference model $R$ consisting of the representations we are investigating
%%Next, we define a reference model $R$, which consists of the representations that are being investigated; these representations are the embeddings of some particular category of word (e.g., the main verbs) from every sentence in $C$.
%%Next, we define a reference model $R$, which consists of the representations that are being investigated; e.g., the embeddings of the main verbs from every sentence in $C$.
Then, we define two hypothesis models, $M_{Hyp_1}$ and $M_{Hyp_2}$. These hypothesis models instantiate hypotheses about the representational geometry of the contextualized word embeddings (Figure \ref{fig:summaryfig}, Step~2). We then draw a sample $c$ of $n$ sentences from our corpus, and calculate the $n \times n$ representational geometries $g$ of each model by applying a dissimilarity metric $D$ to the relevant word embeddings from this sample (Figure \ref{fig:summaryfig}, Step~3). $D(M, c)$ finds the dissimilarity between the representations generated by model $M$ for each pair of sentences in the sample $c$.
\begin{align}
    g_{Ref} &= D(M_{Ref}, c) \\
    g_{Hyp_1} &= D(M_{Hyp_1}, c) \\
    g_{Hyp_2} &= D(M_{Hyp_2}, c)
\end{align}

%\begin{equation}
%    g_{Reference} = D(R, c)
%\end{equation}
%\begin{equation}
%    g_{Hyp1} = D(H_1, c)
%\end{equation}
%\begin{equation}
%    g_{Hyp2} = D(H_2, c)
%\end{equation}
\noindent
Finally, we calculate the similarity $s$ between the representational geometries of our hypothesis models and our reference model using a similarity metric $sim$. Because the $g$ matrices are symmetric, $sim$ only operates on the upper triangle of each $g$ matrix.
\begin{align}
    s_{Hyp_1} &= sim(g_{Ref}, g_{Hyp_1})\\
    s_{Hyp_2} &= sim(g_{Ref}, g_{Hyp_2})
\end{align}

We then repeat the process on $m$ samples from our corpus in order to create two $m$-length vectors of representational similarities, $S_{Hyp_1}$ and $S_{Hyp_2}$ (Figure \ref{fig:summaryfig}, Step~4). Finally, we apply a nonparametric sign test to the difference of these vectors, $S_{Hyp_1} - S_{Hyp_2}$, to test whether there is a consistent difference between measurements of $s_{Hyp_1}$ and $s_{Hyp_2}$. In the following section, we will walk through this approach with a concrete example.

\section{Experiment 1: Subject-Sensitivity of Verb Embeddings}
Most verbs in English require arguments, such as subjects and direct objects. %which are nouns that contribute to the meaning of a verb. 
Here we focus on subjects, which are required by nearly all verbs. 
Specifically, we consider whether the contextualized word embedding of the main verb encodes the subject of a sentence to a greater degree than it encodes other nouns in the sentence that are not arguments of the verb.
%Specifically, we consider whether the subject of a sentence is encoded in the contextualized word embedding of the main verb to a greater degree than other nouns in the sentence that are not arguments. 
In particular, we study the verbs \textit{is} and \textit{are}.

%Because a copula contributes only a grammatical connection without contributing any semantic content, its embedding is likely to be heavily influenced by the relationship that it establishes between the subject and the subject complement, and less heavily influenced by the copula itself. Thus, we would expect the contextualized word embedding of a copula to encode its arguments perhaps even more than an ordinary verb would. 

\paragraph{Corpus Generation:} We generate two corpora using a probabilistic context free grammar (PCFG): the Prepositional Phrase Corpus, containing sentences of the form \ref{copulaPPTemplate}, and the Relative Clause Corpus, containing sentences of the form \ref{copulatRCTemplate}. 

\ex. \a. 
The \textbf{[NOUN$_1$]} \textbf{[PREPOSITION]} the \textbf{[NOUN$_2$]} \textbf{[is/are]} \textbf{[ADJECTIVE]}. \label{copulaPPTemplate}
\b. The \textbf{[NOUN$_1$]} that \textbf{[VERB]} the \textbf{[NOUN$_2$]} \textbf{[is/are]} \textbf{[ADJECTIVE]}. \label{copulatRCTemplate}

%We then exclude sentences where NOUN$_1$ and NOUN$_2$ are identical. 
The following is an example sentence from the Prepositional Phrase Corpus:
\ex. The doctors by the cars are ugly. \label{doctors}

Importantly, any of the nouns in our grammar can occupy the [NOUN$_1$] position or the [NOUN$_2$] position. For example, both sentence \ref{doctors} and the following sentence appear in the Prepositional Phrase Corpus:
\ex. The cars by the doctors are ugly.

Additionally, the two nouns in any given sentence are of the same syntactic number (singular vs. plural) in order to eliminate cues from number features. 
We generated 2,000 sentences for each corpus and then removed repeated sentences and sentences for which NOUN$_1$ and NOUN$_2$ are identical,
%We generated 2,000 sentences for each corpus and then removed examples that did not meet the criteria specified above, 
leaving 1,863 sentences in the Prepositional Phrase corpus and 1,869 in the Relative Clause Corpus. Finally, we chose the vocabulary such that every sentence is semantically plausible.

\paragraph{Representational Models:} In this experiment, we analyze the embeddings of verbs. Thus, our \textit{reference model} is the set of contextualized word embeddings of the verb for each sentence, taken from the last layer of BERT  \cite{devlin2019bert}.\footnote{We use representations from the last layer because \newcite{ethayarajh2019contextual} found that BERT representations are more context-specific in higher layers than lower layers. Because we are looking for the effects of contextualization, we study the layer that is most contextualized.} This study uses the BERT-base-uncased variant of BERT, which creates 768-dimensional embeddings. We compare the representational geometry of these embeddings to that of three hypothesis models. These models instantiate the following hypotheses:
\begin{itemize}
    \item[] \textbf{Subject Hypothesis:} The representational geometry of the contextualized embeddings of verbs reflects information about their subjects.
    \item[] \textbf{Non-Argument Hypothesis:} The representational geometry of the contextualized embeddings of verbs reflects information about nouns that are not arguments of the verb. These non-argument nouns are either the objects of prepositions (in the Prepositional Phrase Corpus) or are the nouns within relative clauses (in the Relative Clause Corpus).
    \item[] \textbf{Null Hypothesis:} The representational geometry of the contextualized embeddings of verbs does not reflect information about subjects or non-argument nouns. 
\end{itemize}

\noindent
We create the Subject Hypothesis Model by taking the 300-dimensional (noncontextualized) GloVe embedding \cite{pennington2014glove} of the verb and concatenating it with the 300-dimensional GloVe embedding of the subject noun. We use GloVe embeddings that are pretrained on the Wikipedia~2014~+~Gigaword~5 corpus. The Non-Argument Hypothesis Model concatenates the GloVe embeddings of the verb and non-argument noun. The Null Hypothesis Model concatenates the GloVe embedding of the verb and the embedding of a random noun from our grammar that does not appear in the sentence.\footnote{This means that the Null Hypothesis Model represents information about the verb and \textit{some} noun. If the contextualized word embedding only encoded the verb and the fact that it refers to a noun (without encoding anything specific about that noun), 
%then we would expect the Hypothesis Models and Null Hypothesis Model to achieve approximately equal representational similarity.
then we would expect the Null Hypothesis Model to achieve approximately equal representational similarity as the other Hypothesis Models.
 } We assume that, if BERT's representations of the verbs in our corpora encode the verbs' subjects more saliently than non-argument nouns, then the representational geometry of BERT's representations will be more similar to the representational geometry of the hypothesis model containing only the subject and verb than to the representational geometry of the hypothesis model containing only the non-argument noun and verb.

Note that GloVe embeddings play no role in BERT, and are likely to be very different from BERT embeddings. One clear difference is that they are of different dimensionalities. However, such differences are immaterial in applying RSA, which illustrates one particular strength of this approach: All that is required is for the hypothesis models to have the hypothesized \textit{representational geometry} (which these concatenated GloVe embeddings do), allowing us to abstract away from superficial differences between the models.

This method of creating hypothesis models has two advantages over other plausible approaches. 
First, it is very likely that BERT's embedding of the verb will be strongly influenced by the identity of the verb (\textit{is} vs. \textit{are}). Because we are just interested in the effect of \textit{context} on the representation, we want to control for the effect of the verb's identity on the representational geometries of our hypothesis models.%\footnote{In this specific case, the representational geometry of our hypothesis models will also be impacted by the differences between the (non-contextualized) representations of \textit{is} and \textit{are}.} 
The approach we have chosen allows us to control for this factor by including each verb's GloVe embedding in each hypothesis model.
%; the approach we have chosen allows us to control 
%First, it allows us to control for the similarities between the two verbs represented by the contextualized embeddings. It is very likely that BERT's embedding of the verb will be strongly influenced by the identity of the verb (\textit{is} vs. \textit{are}). Because we are just interested in the effect of \textit{context} on the representation, we want to control for the effect of the verb on the representational geometries of our hypothesis models.\footnote{In this specific case, the representational geometry of our hypothesis models will also be impacted by the differences between the representations of \textit{is} and \textit{are}.} 
An additional advantage of our approach is that the use of GloVe embeddings allows our similarity measures to be more granular than other plausible approaches (such as one-hot encodings of the verb and relevant noun) would allow. 

Finally, we note that these hypothesis models will likely distort the effect of context compared to the BERT verb representation. By construction, 50\% of each vector in all three hypothesis models consists of a noun and 50\% of the vector consists of the verb itself. It would be surprising to learn that the contextualized embedding encoded only this context information, and in exactly this proportion. Thus, we do not expect the absolute fit of the hypothesis models to be very good. However, each hypothesis model makes the same exaggeration, so comparisons between the models are valid even if each model has a poor absolute fit.
%themselves we can make controlled comparisons between the models. compare the models against each other. 
Thus, we consider only the \textit{differences} in representational similarity between hypothesis models, which indicates which hypothesis model provides a better fit to the reference model.

Pitting these models against each other allows us to determine which aspect of context is encoded to a greater degree. If syntactic structure dominates BERT's representations, then we would expect the subject to be encoded to a greater degree than the non-argument noun. However, if BERT has learned to rely on surface heuristics based on linear distance, then we would expect the opposite. In addition to comparing the hypothesis models to each other, we can also compare each hypothesis model to the null model to determine whether the two nouns that appear in the sentence influence the representational geometry of the BERT embedding more than a random noun.

\paragraph{Applying RSA:}
We now perform RSA on our models, as described in Section \ref{RSAdesc}. We specify the sample size $n = 200$, the number of samples $m = 100$, the dissimilarity metric $D =  1 - Spearman's\:\rho$  and  similarity metric $sim = Spearman's\: \rho$. 

\newcite{zhelezniak2019correlation} show that Spearman's $\rho$ is the most appropriate measurement of (dis)similarity for GloVe embeddings, as these embeddings violate the assumptions underlying other common metrics. We perform a similar analysis to show that it is also the most appropriate dissimilarity measurement for BERT embeddings (See Appendix~B). Finally, we compare representational geometries using Spearman's $\rho$ because it is robust and makes few assumptions \cite{diedrichsen2017representational}.

\begin{figure}
\centering
\begin{subfigure}{.4\textwidth}
  \centering
  \includegraphics[width=\linewidth]{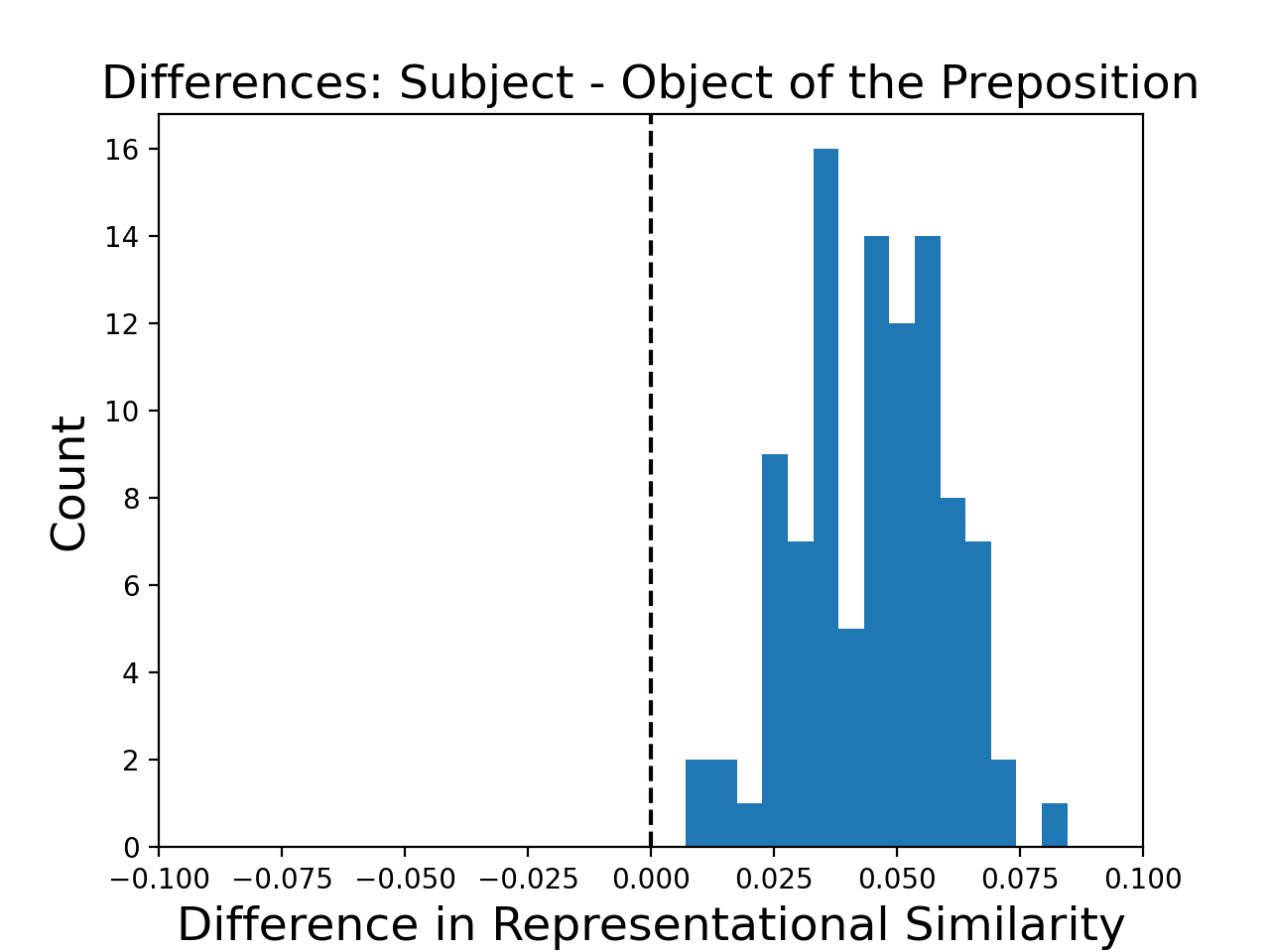}
  \caption{Prepositional Phrase Distribution}
  \label{fig:subj1dist}
\end{subfigure}\hspace{.1\textwidth}
\begin{subfigure}{.4\textwidth}
  \centering
  \includegraphics[width=\linewidth]{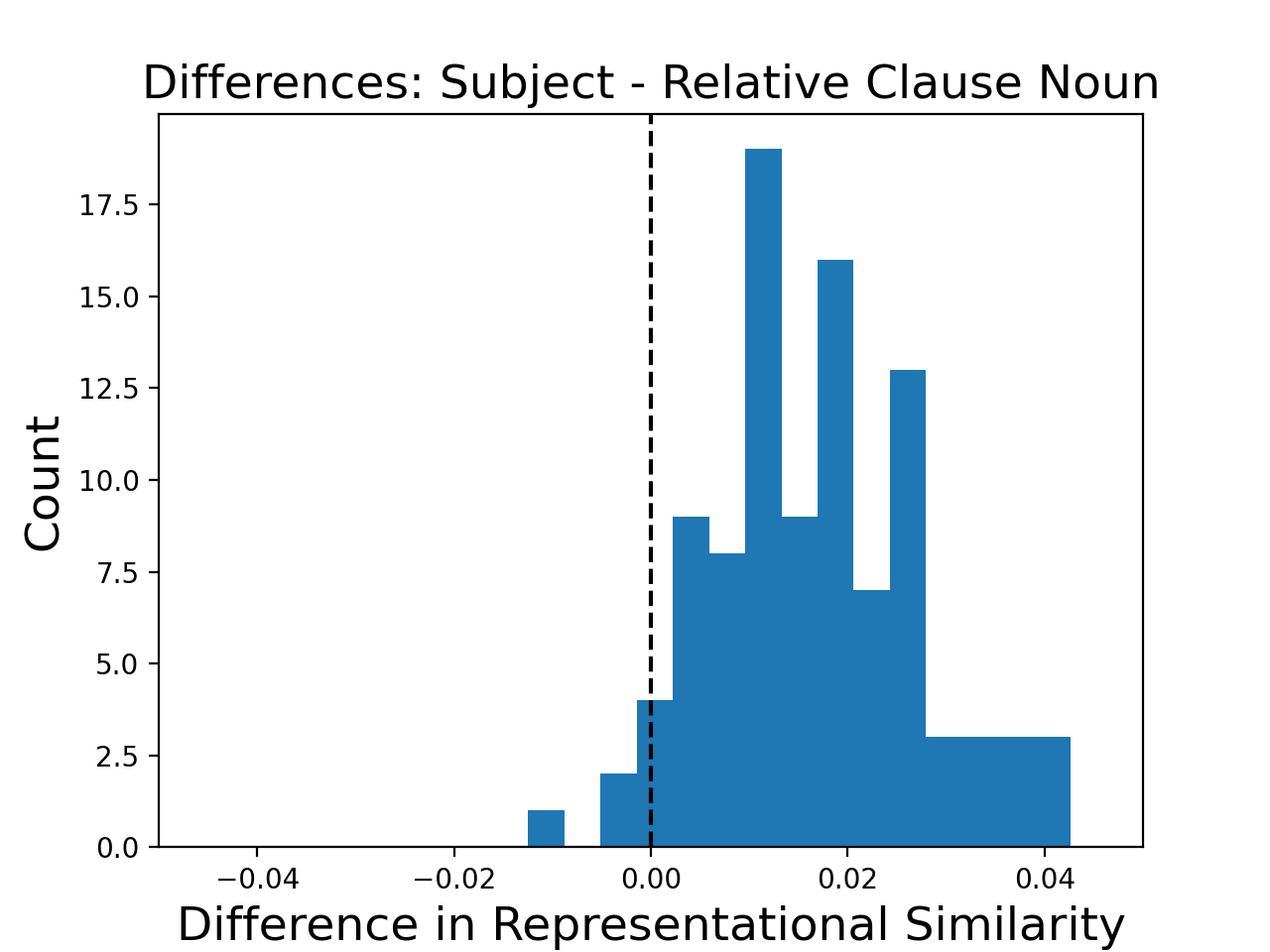}
  \caption{Relative Clause Distribution}
  \label{fig:subj2dist}
\end{subfigure}
\caption{Distributions of differences of representational similarity for Experiment 1. Values greater than 0 denote that the reference model exhibits greater representational similarity to the Subject Hypothesis Model than the Non-Argument Hypothesis Model. For both sentence structures, the Subject Model consistently yields higher representational similarity ($p < .001$).}
\label{fig:subjDist}
\end{figure}

\paragraph{Results:} We summarize our results in Table~\ref{tab:exp1_res}. For both corpora, both the Subject and Non-Argument Hypothesis Models exhibit significantly greater representational similarity to the Reference Model than the Null Hypothesis Model does. This shows that the contextualized representations of verbs encode both of the noun categories in the sentence. Furthermore, these two syntactic categories are not encoded to the same degree, as the Subject Model reliably exhibits greater representational similarity to the Reference Model than the Non-Argument Model does for both corpora (Figure~\ref{fig:subjDist}).
%: in almost all.
%See Figure~\ref{fig:subj1dist} for the distribution of differences for the Prepositional Phrase Corpus and Figure~\ref{fig:subj2dist} for the distribution of differences for the Relative Clause Corpus, and note that the vast majority of differences are positive. This indicates that the Subject Model has greater representational similarity to the reference model. 
From this, we can infer that BERT's embeddings of these verbs encode the verbs' subjects to a greater degree than they encode the non-argument nouns, despite the fact that the non-argument nouns exhibit much less surface-level distance from the verb. 
%This aligns 
%both with the linguistic assertion that arguments contribute meaning to verbs, and 
%with the observation from several other authors that BERT is sensitive to linguistic dependencies, and does not solely rely on surface structure when incorporating contextual information into its embeddings \cite{goldberg2019assessing,klafka-ettinger-2020-spying}.
This result corroborates other evidence---from behavioral evaluations \cite{goldberg2019assessing}, analyses of attention heads \cite{clark2019does}, and probing classifier tests \cite{klafka-ettinger-2020-spying}---that BERT is sensitive to subject-verb dependencies.

 \begin{table}
\centering
\begin{minipage}{.45\linewidth}
\centering
\resizebox{\linewidth}{!}{
\begin{tabular}{l c c c} 
 \toprule
  & Subject & Non-Argument & Null \\ [0.5ex] 
 \midrule
  Prepositional Phrases & \textbf{.524} & .478 &  .440 \\ 
  Relative Clauses & \textbf{.667} & .651 &  .600 \\ 
 \bottomrule
\end{tabular}}
\subcaption{Representational similarities from Experiment 1.}
\label{tab:exp1_res}
\end{minipage}\hfill
\begin{minipage}{.45\linewidth}
\centering
\resizebox{\linewidth}{!}{
\begin{tabular}{l c c c} 
 \toprule
  & Antecedent & Non-Antecedent & Null \\ [0.5ex] 
 \midrule
  Reflexive & \textbf{.717} & .696 & .663 \\ 
  Pronominal & \textbf{.726} & .715 & .691 \\ 
 \bottomrule
\end{tabular}}
\subcaption{Representational similarities from Experiment 2.}\label{tab:exp2_res}
\label{tab:exp2_res}
\end{minipage}
\caption{Representational similarities from Experiments 1 and 2. All differences across columns are statistically significant ($p < .001$). We do not compare across rows because differences in sentence structure make it difficult to compare these numbers.}
\end{table}

% \begin{minipage}{.45\linewidth}
% \centering
% \resizebox{\linewidth}{!}{
% \begin{tabular}{l c c c} 
%  \toprule
%   & Antecedent & Non-Antecedent & Null \\ [0.5ex] 
%  \midrule
%   Anaphors & \textbf{.701} & .676 & .647\\
%   Pronominals & \textbf{.706} & .694 & .669 \\
%  \bottomrule
% \end{tabular}}
% \caption{Results of Experiment 2. All differences are statistically significant, $p < .001$}\label{tab:exp2_res}
% \end{minipage}

% \begin{table}[h!]
% \begin{subfigure}{\columnwidth}
% \centering
% \resizebox{\textwidth}{!}{
% \begin{tabular}{l c c c} 
%  \toprule
%   & Subject & Non-Argument & Null \\ [0.5ex] 
%  \midrule
%   Prepositional Phrases & \textbf{.522} & .496 &  .449 \\ 
%   Relative Clauses & \textbf{.657} & .626 &  .584 \\ 

%  \bottomrule
% \end{tabular}}
% \end{subfigure}%
% \caption{Results of Experiment 1. All differences are statistically significant, $p < .001$}\label{tab:exp1_res}
% \end{table}

\section{Experiment 2: Pronoun Coreference}

Like the meanings of verbs, the meanings of pronouns can also be affected by context. Pronouns typically refer to some contextually-understood noun, which is called its \textit{antecedent}. For example, in \textit{I love New York - it's my favorite city}, \textit{it} refers to \textit{New York}; whereas in \textit{The Death Star is a threat to the galaxy, so it must be destroyed}, \textit{it} refers to \textit{the Death Star}.
Here we investigate whether the contextualized representations of pronouns encode information about the pronouns' antecedents. 

We focus on two types of pronouns, \textit{reflexives} and \textit{pronominals}. Reflexives must refer to a locally-occurring (i.e., in the same clause) noun. Pronominals must \textit{not} refer to a locally-occurring noun. Sentence \ref{reflexive} contains a reflexive (\textit{himself}), which refers to the noun \textit{politician}, while sentence \ref{pronominal} contains a pronominal (\textit{him}), referring to the noun \textit{person}. Note that, in both cases, the pronoun cannot refer to the other noun (underlined).

\ex. \a. The \underline{person} believes that the \textbf{politician} loves \textbf{himself}. \label{reflexive}
\b. The \textbf{person} believes that the \underline{politician} loves \textbf{him}. \label{pronominal}

We study both types of pronoun separately. We thus generate one corpus containing sentences of the same form as Example~\ref{reflexive}, and another containing sentences of the same form as Example~\ref{pronominal}, using a probabilistic context-free grammar. We exclude any sentence where both nouns are the same. This yields 1,828 valid sentences in our reflexive corpus, and 1,826 valid sentences in our pronominal corpus. 

For both corpora, we consider the same reference model: the set of BERT embeddings of the pronouns. We also consider two hypothesis models: the Antecedent Model concatenates the GloVe embedding of the pronoun with the GloVe embedding of its antecedent for every sentence. Note that the antecedent is the first noun in the sentence for the pronominal corpus, and the second noun in the sentence for the reflexive corpus (bolded in Example~\ref{reflexive} and Example~\ref{pronominal}). The Non-Antecedent Model concatenates the GloVe embedding of the pronoun with the GloVe embedding of the non-antecedent noun. The non-antecedent noun is the first noun in the sentence for the reflexive corpus, and the second noun in the sentence for the pronominal corpus (underlined in Example~\ref{reflexive} and Example~\ref{pronominal}). We also include a Null Hypothesis Model, which concatenates the GloVe embedding of the pronoun with the embedding of a random noun from our grammar that does not appear in the sentence.
These hypothesis models allow us to test the hypothesis that the BERT embeddings of pronouns encode their antecedents more than any other noun in the sentence.

\begin{figure}
\centering
\begin{subfigure}{.4\textwidth}
  \centering
  \includegraphics[width=1\linewidth]{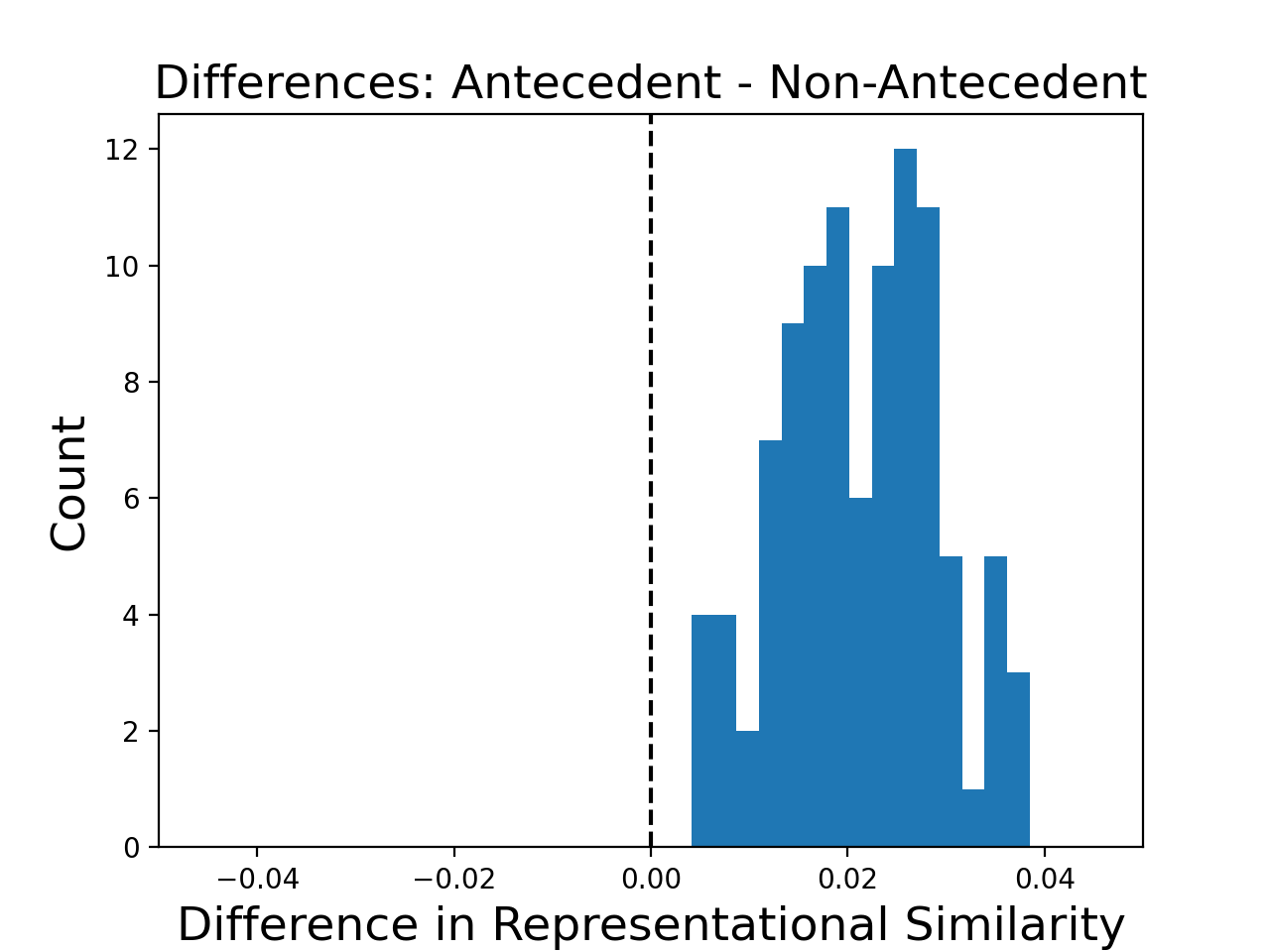}
  \caption{Reflexive Distribution}
  \label{fig:sub1}
\end{subfigure}\hspace{.1\textwidth}
\begin{subfigure}{.4\textwidth}
  \centering
  \includegraphics[width=1\linewidth]{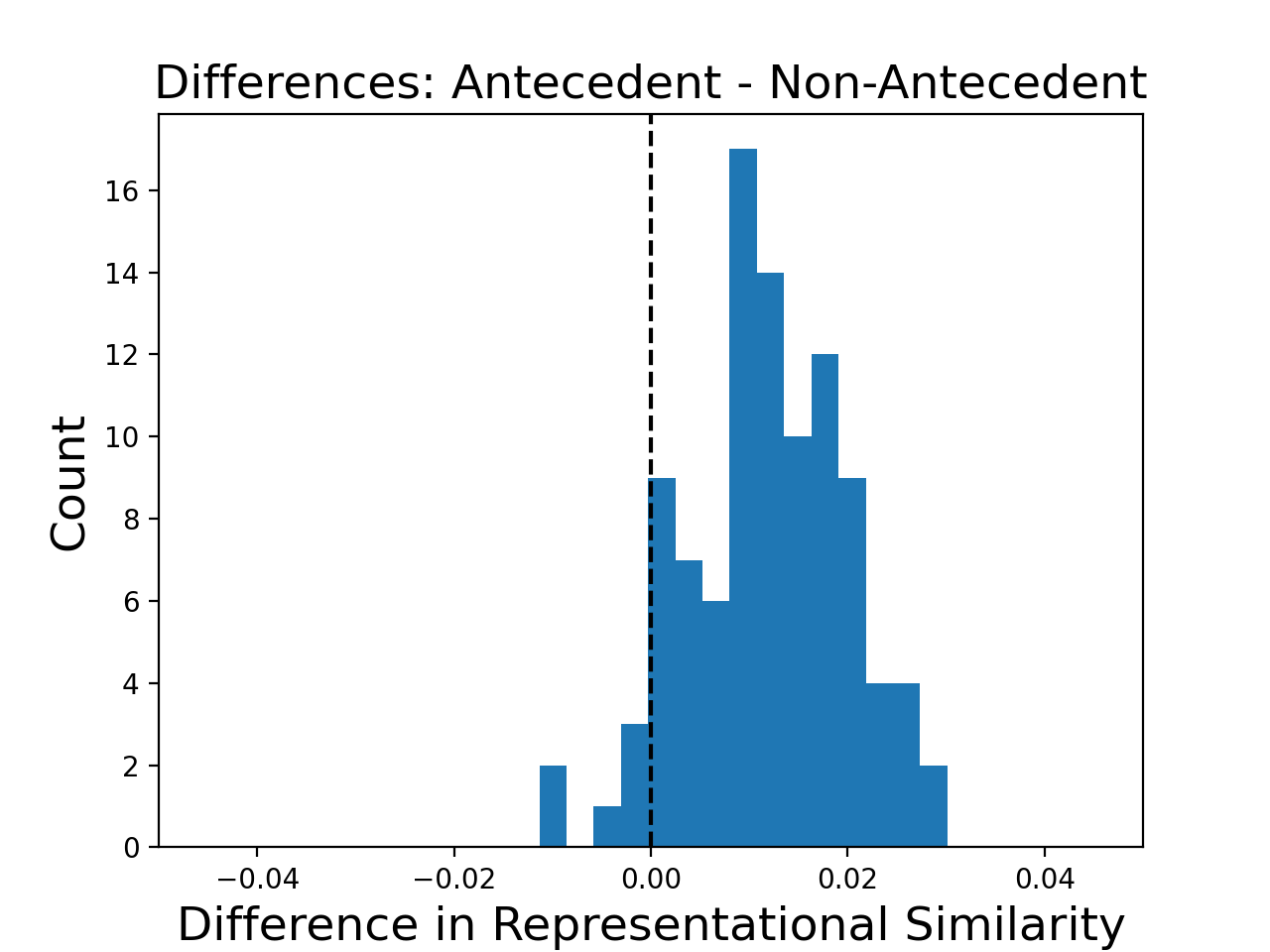}
  \caption{Pronominal Distribution}
  \label{fig:sub2}
\end{subfigure}
\caption{Distributions of differences of representational similarity for Experiment 2. Values greater than 0 denote that the reference model exhibits greater representational similarity to the Antecedent Hypothesis Model than the Non-Antecedent Hypothesis Model. For both sentence structures, the Antecedent Model consistently yields higher representational similarity ($p < .001$).}
\label{fig:corefDist}
\end{figure}

\paragraph{Results:}
We find that, for both corpora, the representational geometry of the set of BERT embeddings of the pronouns is significantly more similar to the hypothesis model that represents antecedent nouns than the hypothesis model that represents non-antecedent nouns (see Table \ref{tab:exp2_res}, as well as Figure~\ref{fig:corefDist}). %For pronominals, the antecedent is the noun that is linearly farther away from the pronoun. Additionally, 
This experiment controls for the absolute and relative linear positions of the words, as the linear position of the antecedent and the non-antecedent swap when the pronoun is a reflexive as opposed to a pronominal.
%this experiment controls for the absolute linear position of the words, as the linear position of the antecedent and the non-antecedent swap when the pronoun is a reflexive as opposed to a pronominal. 
Thus, the representational geometry of BERT pronoun embeddings is also sensitive to syntactic dependencies, as opposed to strictly surface-level cues. 
BERT's sensitivity to the relationship between pronouns and their antecedents---which was also observed by \newcite{clark2019does} through analysis of BERT's attention heads---may explain BERT's strong performance on coreference resolution, a task that relies on identifying pronoun-antecedent relationships \cite{joshi-etal-2019-bert}.

% \begin{table}[h!]
% \begin{subfigure}{\columnwidth}
% \centering
% \resizebox{\textwidth}{!}{
% \begin{tabular}{l c c c} 
%  \toprule
%   & Antecedent & Non-Antecedent & Null \\ [0.5ex] 
%  \midrule
%   Anaphors & \textbf{.701} & .676 & .647\\
%   Pronominals & \textbf{.706} & .694 & .669 \\
%  \bottomrule
% \end{tabular}}
% \end{subfigure}%
% \caption{Results of Experiment 2. All differences are statistically significant, $p < .001$}\label{tab:exp2_res}
% \end{table}

\section{Experiment 3: Heads of Sentences}

Our approach can be applied to any type of representation that incorporates information from multiple words. So far we have applied it to contextualized word representations, but it can also be applied to full-sentence representations; here we give an example of such a usage inspired by dependency parsing. 
%See Figure~\ref{DepParse} for an example of a dependency parse. 
%In standard approaches to dependency parsing \cite{de-marneffe-etal-2006-generating,nivre2016universal}, the main verb of a sentence acts as the head of the entire sentence, as in Figure~\ref{DepParse}. 
%The motivation behind this is that sentences typically describe events. Verbs describe that event and also establish various roles that must be filled by nouns in the sentence. For example, the sentence in Figure~\ref{DepParse} describes a giving event, which necessitates 3 roles: the giver, the receiver, and the thing that is given.
\begin{figure}
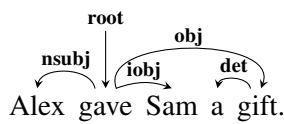

\centering
\begin{dependency}[theme = simple,
 label style={font=\bfseries}]
   \begin{deptext}[]
      Alex \& gave \& Sam \& a \& gift.\\
   \end{deptext}
   \deproot[edge height=1cm]{2}{root}
   \depedge{2}{1}{nsubj}
   %\depedge[edge start x offset=-6pt]{2}{5}{}
   \depedge[arc angle=20]{2}{3}{iobj}
   \depedge[arc angle=110]{2}{5}{obj}
   \depedge{5}{4}{det}

   %\depedge[arc angle=50]{7}{6}{}
\end{dependency}
    \caption{Example dependency parse}
   \label{DepParse}
\end{figure}

In standard approaches to dependency parsing \cite{de-marneffe-etal-2006-generating,nivre2016universal}, the main verb of a sentence acts as the head of the entire sentence, as in Figure~\ref{DepParse}. 
We study whether verbs exhibit the same primacy in BERT embeddings as they do in dependency parses. To do so, we use the embedding of the [CLS] token as an embedding of the full sentence, as is standard when a sentence embedding is required from BERT. We create four corpora, each containing a different type of sentence structure. These corpora are denoted the Intransitive Corpus, Intransitive + Adjective Corpus, Transitive Corpus, and Transitive + Adjective Corpus, respectively. An example from each corpus is shown in Table~\ref{Exp3Corps}.

\begin{table}
\centering
\resizebox{.65\textwidth}{!}{
\begin{tabular}{l l} 
 \toprule
 Corpus & Example \\ [0.5ex] 
 \midrule
  Intransitive & The painter swims. \\ 
  Intransitive + Adjective & The happy politician talks.  \\ 
  Transitive & The person moves a lamp. \\ 
  Transitive + Adjective & A scary lawyer likes the red chair. \\ 
 \bottomrule
\end{tabular}}
\caption{Corpora used in Experiment 3.}
\label{Exp3Corps}
\end{table}

For each corpus, we use the set of BERT embeddings for each sentence's [CLS] token as our reference model. We then consider 
%one hypothesis model for each content word (i.e. each noun, adjective, or verb) in the sentence, 
one hypothesis model for each type of content word (i.e., verb, subject noun, direct object noun, adjective modifying the subject, or adjective modifying the direct object),
where each hypothesis model consists of the GloVe embeddings for the instances of the relevant type of content word. We specify 10 unique words that can appear in the position of every content word. For example, in the Transitive + Adjective Corpus the first and second adjectives are randomly selected from separate, non-overlapping vocabularies of 10 words, as are the two nouns. Thus, the number of words that can fill each slot is matched across conditions. We also include a Null Hypothesis Model, which consists of a GloVe embedding for a verb in our vocabulary that does not appear in the sentence.

These hypothesis models allow us to determine which individual content word is encoded to the greatest degree by the full sentence embedding. Standard  dependency-parsing frameworks suggest that the main verb is the most salient word in a sentence. If BERT's full sentence embeddings reflect this intuition, then we would expect the representational geometry of the Verb Model to exhibit the greatest similarity to the representational geometry of our reference model.

 When we perform RSA on the models using the Intransitive Corpus, we use a sample size $n$ of 50, as there are only 200 sentences in that corpus. The remaining corpora are larger, allowing for sample sizes $n$ of 200 (the Intransitive + Adjective Corpus has 1,273 sentences, Transitive Corpus has 1,572 sentences, and Transitive + Adjective Corpus has 1,995 sentences, after generating 2,000 sentences from a PCFG for each one and then excluding duplicate sentences).

\paragraph{Results:}
Across all sentence types, the Verb Model exhibits the greatest representational similarity to our reference model (Table \ref{tab:exp3_res}). In addition, the Verb Model exhibits a significantly higher representational similarity to the Reference Model than the Null Hypothesis Model does. Thus, the type of content word that is encoded to the greatest degree in the representational geometry of BERT's full sentence embeddings is the verb, a result that aligns with the primacy of verbs in standard dependency parsing formalisms.

%Though the primacy of the verb is in line with standard dependency frameworks, 
We also find that, for both the Transitive Corpus and Transitive + Adjective Corpus, the Object Model exhibits a greater representational similarity to the reference model than the Subject Model does. We do not have an explanation for why the direct object would play a greater role in the sentence representation than the subject does.

%It is also notable that, for both the Transitive Corpus and Transitive + Adjective Corpus, the Object Model exhibits a greater representational similarity to the reference model than the Subject Model. Though the dependency framework asserts that both arguments are of equal importance in a sentence, other linguistic work suggests that objects are more important to the meaning of verbs than subjects \cite{larson1988double,marantz1997no}. These results support this proposition.

\begin{table}[h!]
\centering
\begin{subfigure}{.75\linewidth}
\resizebox{\textwidth}{!}{
\begin{tabular}{l c c c c c c} 
 \toprule
  & Verb & Subject & Subj. Adj. &  Object & Obj. Adj. &  Null  \\ [0.5ex] 
 \midrule
  Intransitive & \textbf{.221} & .104 & - & - & - & -.034 \\ 
  Intransitive + Adj & \textbf{.218} & .113 & .148 & - & - & .010 \\ 
  Transitive & \textbf{.301} & .113 & - & .185 & - & .017 \\ 
  Transitive + Adj & \textbf{.313} & .094 & .083 & .132 & .055 & .014 \\ 

 \bottomrule
\end{tabular}}
\end{subfigure}%
\caption{Results of Experiment 3. All differences between the Verb Model and any other model are statistically significant, as are both differences between the Subject Model and Object Model ($p < .001$). `-' denotes that the corpus did not contain content words of that type.} \label{tab:exp3_res}
\end{table}

\section{Comparisons to Other Approaches} \label{comp}

One popular approach for analyzing neural network models is the \textit{behavioral} approach, in which a model's performance is evaluated on some challenge set designed to highlight particular linguistic phenomena. 
These methods have been used to determine whether various language models track dependencies based on syntactic number \cite{linzen2016assessing,gulordava-etal-2018-colorless,goldberg2019assessing}. Additionally, natural language inference tasks directly test a model's ability to infer semantic relationships \cite{mccoy-etal-2019-right,bowman-etal-2015-large}. For more examples, see \newcite{belinkov2019analysis}.
Behavioral methods are well-suited for holistic questions about a model's overall handling of language, but here our goal is to analyze specific internal representations independently of behavior. 
For this question, the behavioral approach is less well-suited because it can only give an indirect window into the structure of the representational space; analyzing the internal representations themselves is much more direct.

Another popular approach, the \textit{diagnostic model} approach, permits analyses that directly investigate internal representations.
Diagnostic models are simple (usually linear) classification or regression models that take in activation patterns and predict some information of interest. If the diagnostic model performs well, then the activation pattern is typically said to encode this information. Prior work has used this method to determine that BERT's representations are sensitive to the hierarchical structure of language \cite{lin-etal-2019-open}, contain information relevant to a variety of tagging tasks \cite{liu2019linguistic}, contain semantic information \cite{tenney2019you}, and contain much other linguistically-useful information \cite{rogers2020primer,belinkov2019analysis}. 

These usages of diagnostic models are generally focused on testing whether an embedding \textit{does or does not} represent certain information, but we instead focus on the \textit{degree} to which the embedding represents information from particular words, and standard diagnostic-model approaches make it difficult to obtain relative results of this sort, because it is possible that information from all words in the sentence is represented in the embedding to some degree. This claim is supported by recent work, which finds that several salient linguistic features can be recovered from \textit{every} word in a sentence using diagnostic classifiers \cite{klafka-ettinger-2020-spying}.
Thus, it seems that a diagnostic classifier approach may be insufficient for answering the questions that we address here. Indeed, we attempted to devise a diagnostic model approach and applied it to our pronoun coreference and verb subject-sensitivity studies, but found little success (See Appendix~A). 
A modification to the diagnostic model approach which makes this approach more amenable to illuminating relative strengths of encoding is minimum description length (MDL) probing, introduced in the concurrent work of \newcite{voita2020information}.
MDL probing characterizes how regularly specific information is encoded in vectors, while our approach investigates the representational geometry of a set of encodings. These concepts of regularity and representational geometry are likely related, but we leave for future work an investigation of the precise relationship between them.

One final approach that has been used to study whether linguistic dependencies are captured in BERT is to analyze whether specific attention heads track particular dependency relationships \cite{clark2019does,htut2019attention}. Though this approach enables analysis of the same linguistic phenomena that we study, it does not directly demonstrate whether or how this information is encoded in BERT's vector representations. Instead, it illustrates which inputs are most salient in generating these vector representations, without revealing what information from those inputs is encoded in the final representation. In addition, because BERT uses many attention heads, the behavior of any one attention head gives only a partial picture of the inputs for a given vector.

\section{Related Work}
\paragraph{Other Applications of RSA in NLP:}
Previous work has also explored the application of Representational Similarity Analysis to neural networks. Indeed, one of the first applications of this type of analysis was performed on artificial neural systems \cite{laakso2000content}. \newcite{abnar-etal-2019-blackbox} use RSA to characterize how the representational geometries of various models change when given different amounts of context. Other work has used RSA to compare the semantic representations 
of CNN object detection models with that of word vectors \cite{dharmaretnam2018emergence}, to compare the representations of utterances in a spoken-word encoder to the representations of those same words in the text or visual domain \cite{chrupala-2019-symbolic}, and to compare the representational spaces of two agents in an emergent language game \cite{bouchacourt-baroni-2018-agents}.

\newcite{chrupala2019correlating} also employ RSA to study the correspondence between the representational geometry of various language models (including BERT) and a hypothesis model based on gold syntax trees. They introduce a new technique, $RSA_{regress}$, which merges RSA with a diagnostic model approach. The present work is complementary to their study. Whereas their study compares different neural models against one hypothesis model, our study compares multiple hypothesis models against a single neural model, in order to adjudicate between specific linguistic hypotheses. 

\paragraph{Geometric Analyses of NLP Systems:}
Though we study the representational geometry of contextualized word embeddings, other work has been done to characterize the geometry of the network activations themselves. These are distinct concepts, as our current work can be thought of as analyzing the second-order geometry (comparing the relationships between representations of one model to the relationships between representations of another), while this prior work is analyzing first-order geometry (comparing the representations of one model to the representations of another). This first-order approach has also been used to analyze models' linguistic capabilities \cite{wu2020similarity,reif2019visualizing,kimcompositionality,hewitt2019structural}. Most directly related, \newcite{lin-etal-2019-open} uses a first-order approach to analyze BERT's representation of subject-verb agreement and pronoun coreference.

\section{Conclusion and Future Work} \label{futwork}
%Assessing the linguistic abilities of NLP systems is essential for understanding how to improve them. 
We have introduced a framework for using representational similarity analysis to adjudicate between hypotheses about the representational geometry of a neural network's embeddings.
%We use representational similarity analysis to adjudicate between hypotheses about the representational geometry of a neural network's embeddings.
%the linguistic representations of neural networks. 
We then applied this procedure to BERT embeddings, and demonstrated that they are sensitive to linguistic dependencies. In particular, we showed that BERT's embeddings of pronouns encode the pronouns' antecedents more than they encode other nouns, that the embeddings of verbs encode the verbs' subjects more than they encode non-argument nouns, and that BERT's sentence embeddings most saliently encode the sentences' main verbs, as predicted by standard dependency frameworks. Not only do these studies reveal a sensitivity to linguistic dependencies, two of them (the subject-sensitivity and pronoun-coreference studies) demonstrate that these dependencies are \textit{more} salient than relationships between words at the surface level. 

This framework can enable the investigation of many linguistically-motivated questions.
%allows us to ask a plethora of linguistically-motivated questions. 
As long as hypothesis models can be defined to instantiate the relevant linguistic hypotheses, this framework allows researchers to study which of the hypotheses is closer to the truth. 
Additionally, future work can focus on making more complete hypothesis models. Our studies only required us to account for the \textit{differences} between the representational similarity of different hypothesis models, but not the \textit{magnitude} of that similarity. In future work, we plan to focus on creating more complex hypothesis models that exhibit greater absolute representational similarity to the reference model. %Indeed, it may even be possible to determine the precise proportion of the representational geometry that each word in a sentence accounts for. This could be done by defining hypothesis models that concatenate all words in the sentence together in varying proportions, and analyzing their relative and absolute representational similarities.

\section*{Acknowledgments}

We are grateful to Tal Linzen for helpful comments. Any errors are our own.
This work was supported by the National Science Foundation Graduate Research Fellowship Program under Grant No. 1746891. Any opinions, findings, and conclusions or recommendations expressed in this material are those of the authors and do not necessarily reflect the views of the National Science Foundation.

\bibliographystyle{coling}
\bibliography{coling2020}

\newpage
\appendix
\section*{Appendix A: Diagnostic Classifier Comparison}
\label{diag}

We attempted to investigate the argument-sensitivity of BERT's embeddings of verbs using diagnostic models. To do so, we trained two separate logistic regression classifiers.\footnote{All logistic regression classifiers used the default Scikit-Learn hyperparameters.} The input for both classifiers was the GloVe embedding of any word in the sentence \textit{except the main verb} concatenated with the BERT embedding of the verb. The Subject Classifier was trained to classify whether the GloVe embedding corresponded to the subject of the verb or not. The Non-Argument Noun Classifier was trained to classify whether the GloVe embedding of a word corresponded to the non-argument noun or not. We train these classifiers for both the Prepositional Phrase Corpus and the Relative Clause Corpus. For both classifiers, we use 80\% of the data for training and 20\% for testing. Looking at Table~\ref{tab:verbDiag}, we see that all classifiers failed to do better than majority-class performance.

We performed a similar investigation of pronoun coreference. The input for our classifiers was the GloVe embedding of any word in the sentence \textit{except the pronoun}, concatenated with the BERT embedding of the pronoun. For both the pronominal and reflexive corpora, we created an Antecedent Classifier (which was trained to identify the antecedent of a sentence), and a Non-Antecedent Classifier (which was trained to identify the non-antecedent noun of a sentence). For both classifiers, we use 80\% of the data for training and 20\% for testing. Looking at Table~\ref{tab:corefDiag}, we see that all classifiers failed to do better than majority class performance.\footnote{Note: In the original publication, the Reflexive Non-Antecedent Precision and Recall were incorrectly listed. These are the correct values.}

\begin{table}[h!]
\begin{subfigure}{\columnwidth}
\centering
\resizebox{.7\textwidth}{!}{
\begin{tabular}{l c c c} 
 \toprule
  & Accuracy & Precision & Recall \\ [0.5ex] 
 \midrule
  PP Subject & .832 & .023 &  .003 \\ 
  PP Non-Argument & .831 & .074 &  .013  \\ 
  RC Subject & .866 & 0.0 &  0.0 \\ 
  RC Non-Argument & .870 & .224 &  .037 \\ 

 \bottomrule
\end{tabular}}
\end{subfigure}%
\caption{Results of the Diagnostic Classifier approach to Experiment 1. Note that the majority class accuracy for the prepositional phrase corpus classifiers is .83, as 5 out of 6 tested words in every sentence are not subjects/non-argument nouns. The majority class accuracy for the relative clause corpus classifiers is .87, as 6 out of the 7 tested words in every sentence are not subject/non-argument nouns}\label{tab:verbDiag}
\end{table}

\begin{table}[h!]
\begin{subfigure}{\columnwidth}
\centering
\resizebox{.7\textwidth}{!}{
\begin{tabular}{l c c c} 
 \toprule
  & Accuracy & Precision & Recall \\ [0.5ex] 
 \midrule
  Reflexive Antecedent & .865 & .045 &  .003 \\ 
  Reflexive Non-Antecedent & .875 & .15 &  .018  \\ 
  Pronominal Antecedent & .867 & .042 &  .003 \\ 
  Pronominal Non-Antecedent & .870 & .162 &  .017  \\ 
 \bottomrule
\end{tabular}}
\end{subfigure}%
\caption{Results of the Diagnostic Classifier approach to Experiment 2. Note that the majority class accuracy for the Antecedent and Non-Antecedent classifiers is .87, as 6 out of 7 tested words in every sentence are not antecedents/non-antecedent nouns.}\label{tab:corefDiag}
\end{table}

\section*{Appendix B: Spearman's $\rho$ for Embeddings}
\label{spearmans}

\newcite{zhelezniak2019correlation} show that it is most appropriate to use Spearman's $\rho$ as a (dis)similarity metric when embeddings are non-normal. They go on to show that GloVe embeddings violate the assumption of normality by using Shapiro-Wilk tests and analyzing Q-Q plots of some GloVe embeddings. Here, we show that the same result holds for BERT embeddings, and thus show that it is most correct to use Spearman's $\rho$ (as opposed to cosine similarity or other metrics) as a (dis)similarity metric for BERT embeddings as well.

We analyze our Prepositional Phrase, Relative Clause, Reflexive, and Pronominal Corpora. For these analyses, we only exclude repeated sentences. The resulting Prepositional Phrase Corpus contains 1,935 sentences, the Relative Clause Corpus contains 1,944 sentences, the Reflexive Corpus contains 1,909 sentences, and the Pronominal Corpus contains 1,899 sentences.

Following \newcite{zhelezniak2019correlation}, we treat each embedding as a `sample of observations from a scalar random variable'. First, we Z-normalize all embeddings, such that they have a mean of 0 and standard deviation of 1. We then apply a Shapiro-Wilk test to the BERT embedding corresponding to every word in the corpus (i.e. not the [CLS] or [SEP] token embeddings). We first apply these tests to the full BERT embeddings, and then repeat the procedure on a random subsample (without replacement) of 300 values from each of these embeddings. We subsample in order to show that a large number of embeddings are still found to be non-normal even with a smaller sample size. See Table~\ref{tab:shapirowilk} for the results.

\begin{table}[h!]
\centering
\resizebox{\textwidth}{!}{
\begin{tabular}{l c c c} 
 \toprule
  Corpus & Full Embedding \% Non-Normal & Sampled Embedding \% Non-Normal & \# of Embeddings \\ [0.5ex] 
 \midrule
  Reflexive & 99.98 & 48.24 & 15,272 \\ 
  Pronominal & 100 & 49.01 &  15,192  \\ 
  Prepositional Phrase & 99.92 & 48.23 &  13,545 \\ 
  Relative Clause & 99.89 & 46.59 &  15,552 \\ 
 \bottomrule
\end{tabular}}
\caption{Results from Shapiro-Wilk tests performed on BERT embeddings for every word in each corpus. Embeddings are considered non-normal if the Shapiro-Wilk test returns a P-value less than .05.}\label{tab:shapirowilk}
\end{table}

\begin{figure}
\centering
\begin{subfigure}{.4\textwidth}
  \centering
  \includegraphics[width=\linewidth]{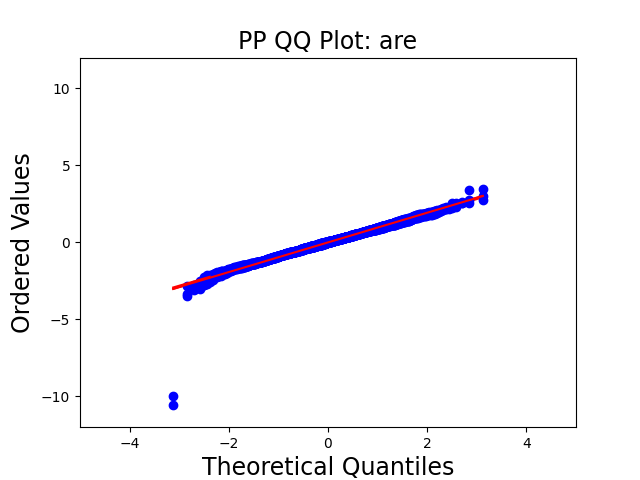}
  \caption{Prepositional Phrase QQ Plot}
  \label{fig:subj1}
\end{subfigure}\hspace{.1\textwidth}
\begin{subfigure}{.4\textwidth}
  \centering
  \includegraphics[width=\linewidth]{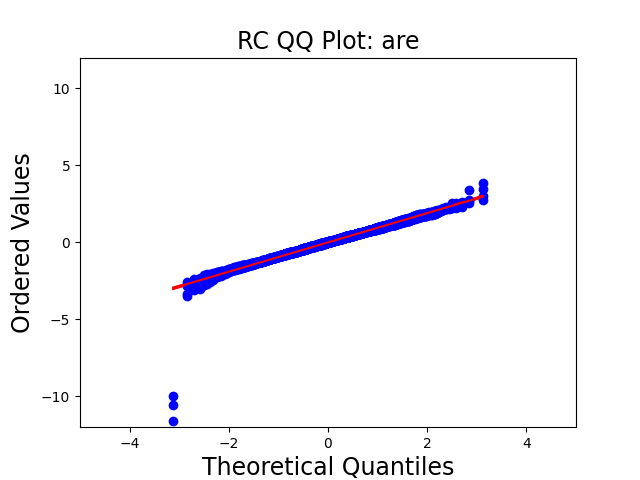}
  \caption{Relative Clause QQ Plot}
  \label{fig:subj2}
\end{subfigure}
\begin{subfigure}{.4\textwidth}
  \centering
  \includegraphics[width=\linewidth]{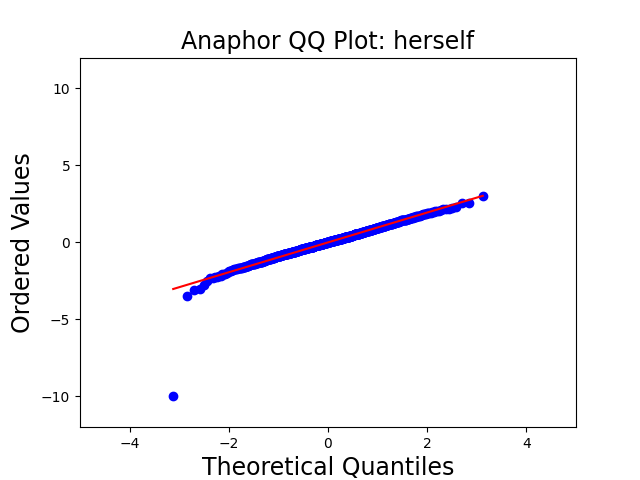}
  \caption{Reflexive QQ Plot}
  \label{fig:subj2}
\end{subfigure}\hspace{.1\textwidth}
\begin{subfigure}{.4\textwidth}
  \centering
  \includegraphics[width=\linewidth]{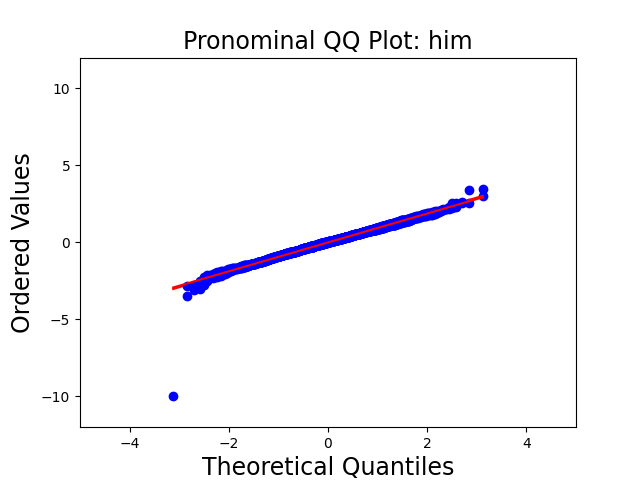}
  \caption{Pronominal QQ Plot}
  \label{fig:subj2}
\end{subfigure}
\caption{QQ Plots for BERT embeddings from each corpus. We choose embeddings that we use to construct our reference models in Experiments 1 and 2. All plots demonstrate the presence of at least one large outlier, and several other smaller outliers.}
\label{fig:QQ_Plots}
\end{figure}

Furthermore, we analyze one BERT embedding for each corpus using Q-Q plots. We see from Figure~\ref{fig:QQ_Plots} that these embeddings contain significant outliers, indicating that they are not normally distributed.

\end{document}